\documentclass[sigconf]{acmart}

\usepackage{multirow}

\usepackage{amssymb}

\AtBeginDocument{%
  }

\copyrightyear{2023}
\acmYear{2023}
\setcopyright{acmlicensed}
\acmConference[MM '23]{Proceedings of the 31st ACM International Conference on Multimedia}{October 29-November 3, 2023}{Ottawa, ON, Canada}
\acmBooktitle{Proceedings of the 31st ACM International Conference on Multimedia (MM '23), October 29-November 3, 2023, Ottawa, ON, Canada}
\acmPrice{15.00}
\acmDOI{10.1145/3581783.3613763}
\acmISBN{979-8-4007-0108-5/23/10}

\begin{document}

\title{Multi-scale Target-Aware Framework for Constrained Image Splicing Detection and Localization}

\author{Yuxuan Tan}
\email{2110436102@email.szu.edu.cn}
\affiliation{%
  \institution{Guangdong Key Laboratory of Intelligent Information Processing, College of Electronics and Information Engineering, Shenzhen University
  \city{Shenzhen}
  \country{China}}
}

\author{Yuanman Li$^{*}$ }
\email{yuanmanli@szu.edu.cn}
\thanks{$^{*}$Corresponding author}
\affiliation{%
  \institution{Guangdong Key Laboratory of Intelligent Information Processing, College of Electronics and Information Engineering, Shenzhen University
  \city{Shenzhen}
  \country{China}}
}

\author{Limin Zeng}
\email{2100432013@email.szu.edu.cn}
\affiliation{%
  \institution{Guangdong Key Laboratory of Intelligent Information Processing, College of Electronics and Information Engineering, Shenzhen University
  \city{Shenzhen}
  \country{China}}
}

\author{Jiaxiong Ye}
\email{yejiaxiong2021@email.szu.edu.cn}
\affiliation{%
  \institution{Guangdong Key Laboratory of Intelligent Information Processing, College of Electronics and Information Engineering, Shenzhen University
  \city{Shenzhen}
  \country{China}}
}

\author{Wei Wang}
\email{ehomewang@ieee.org}
\affiliation{%
  \institution{Department of Engineering, Shenzhen MSU-BIT University
  \city{Shenzhen}
  \country{China}}
}

\author{Xia Li}
\email{lixia@szu.edu.cn}
\affiliation{%
  \institution{Guangdong Key Laboratory of Intelligent Information Processing, College of Electronics and Information Engineering, Shenzhen University
  \city{Shenzhen}
  \country{China}}
}








\renewcommand{\shortauthors}{Yuxuan Tan et al.}

\begin{abstract}
Constrained image splicing detection and localization (CISDL) is a fundamental task of multimedia forensics, which detects splicing operation between two suspected images and localizes the spliced region on both images. Recent works regard it as a deep matching problem and have made significant progress. However, existing frameworks typically perform feature extraction and correlation matching as separate processes, which may hinder the model's ability to learn discriminative features for matching and can be susceptible to interference from ambiguous background pixels.
In this work, we propose a multi-scale target-aware framework to couple feature extraction and correlation matching in a unified pipeline. In contrast to previous methods, we design a target-aware attention mechanism that jointly learns features and performs correlation matching between the probe and donor images. Our approach can effectively promote the collaborative learning of related patches, and perform mutual promotion of feature learning and correlation matching. Additionally, in order to handle scale transformations, we introduce a multi-scale projection method, which can be readily integrated into our target-aware framework that enables the attention process to be conducted between tokens containing information of varying scales. Our experiments demonstrate that our model, which uses a unified pipeline, outperforms state-of-the-art methods on several benchmark datasets and is robust against scale transformations.
\end{abstract}



\maketitle

\section{Introduction}
With the advancement of image editing software and the rapid growth of social networks, digital images can be easily manipulated and widely spread on the Internet. Some maliciously tempered images can be used to fabricate false evidence or manipulate public opinion, leading to negative social influence. Therefore image forensics technology, which aims to verify the authenticity of images and localize the forgery region, has attracted great attention in research \cite{wang2009survey}. 

Conventional image forensics methods investigate a single potential forgery image and attempt to find out the forgery region by analyzing the manipulation traces such as light inconsistency \cite{light}, motion blur \cite{blur}, photo-response noise \cite{photo}, etc. However, various post-processing transformations (e.g., compression, resizing, noise addition, rotation) usually be applied to hide these weak manipulation traces \cite{zhong_end--end_2020,li_fast_2019,cozzolino_efficient_2015,chen_serial_2020}, which significantly reduces the effectiveness of forgery detection algorithms\cite{barni2012hide}. 

\begin{figure}[t!]
	\centering
	\includegraphics[width=0.9\linewidth]{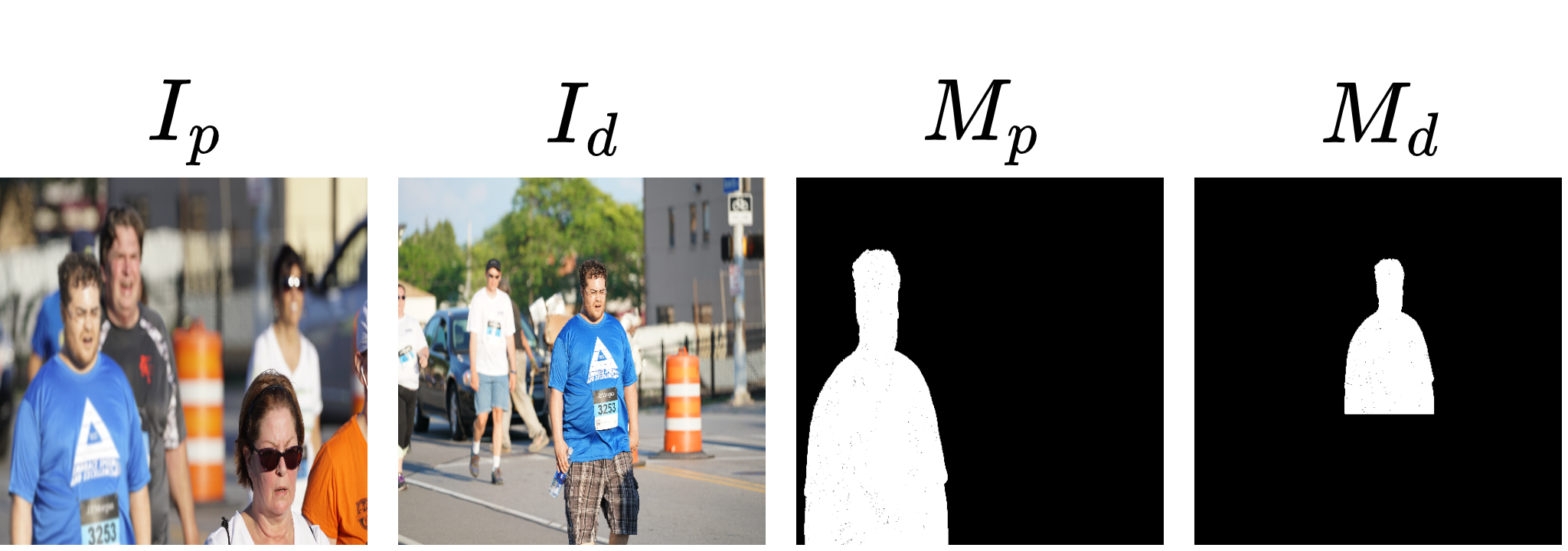}
	\centering
	\caption{Visual presentation of constrained image splicing detection and localization (CISDL). $I_p$ and $I_d$ represent the probe image and the donor image respectively, $M_p$ and $M_d$ are their corresponding masks, where white pixels indicate the spliced region.}
	\label{fig:Fig1}
\end{figure}
In order to promote a new reliable forensics method, Media Forensics Challenge proposed a new forensics task that both detects the forgery region and the corresponding source region. Some researchers refer to it as the constrained image splicing detection and localization (CISDL) problem \cite{liu2019adversarial}. As shown in Fig.\ref{fig:Fig1}, given a probe image $I_p$ and a potential donor image $I_d$, CISDL aims to determine if the probe image $I_p$ contains regions spliced from the donor image $I_d$, and provide two masks indicating the spliced region for both images. Compared to conventional image forgery detection, CISDL takes a pair of images as input, which provides more information for detecting the spliced region. 

\begin{figure}[t!]
	\centering
	\includegraphics[width=0.9\linewidth]{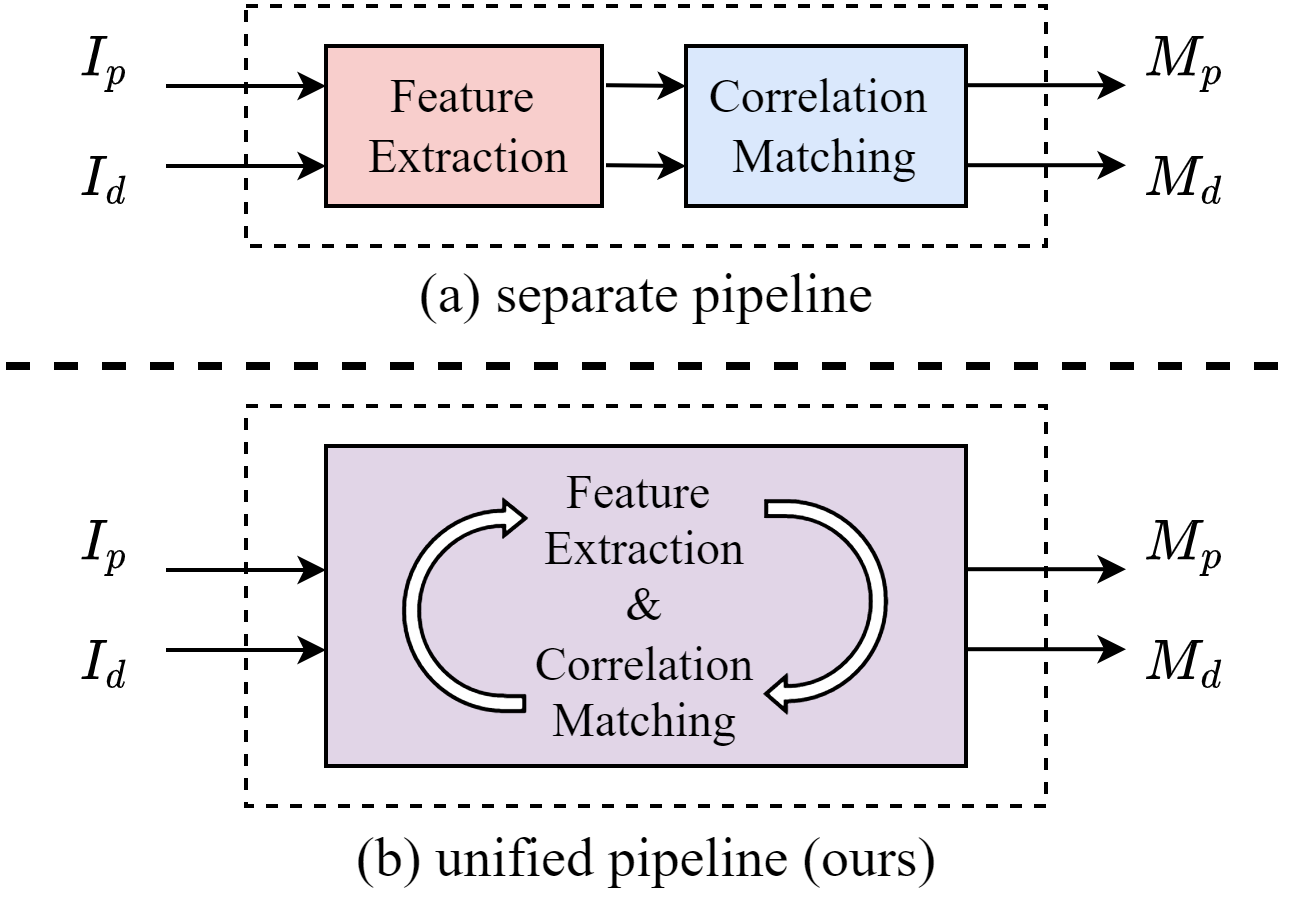}
	\centering
	\caption{Two different pipelines for CISDL.}
	\label{fig:pipeline}
\end{figure}

In recent years, CISDL has garnered growing interest, and many works have been devised \cite{wu2017deep, ye2018feature,liu2019adversarial,liu2020constrained, xu2022scale}. While existing methods for CISDL have made significant progress, their models are often structured as separate pipelines. This can limit the model's ability to learn discriminative features for matching and make it vulnerable to interference from ambiguous background pixels. As shown in Fig. \ref{fig:pipeline} (a), they mainly take CNN as the feature extractor to capture the semantic information of the input images. Afterward, a correlation matching module is designed to compare these high-level features and output correlation score maps to determine which regions in the two input images form matching. These separate pipeline models have certain shortcomings: (1) The feature extraction process of the two images is independent, which can not utilize the correlation of similar image patches between the two images to perceive the target features. (2) For the matching problem, sufficient feature interaction between the two images is crucial. However, these separate pipeline models only perform interaction in the last single correlation matching module, leading to insufficient information interaction between two images. 

Inspired by \cite{cui2022mixformer, chen2022backbone}, we design a unified pipeline for CISDL as shown in Fig.\ref{fig:pipeline}(b), which can greatly address those problems mentioned above. Different from the previous separate pipeline, our unified pipeline framework jointly conducts feature extraction and correlation matching, which effectively promotes collaborative learning of related patches in the probe image and the donor image. This design can perform mutual promotion for feature extraction and correlation matching. With the help of correlation matching, our unified pipeline model can greatly perceive the related region and suppress the effects from the background and other unrelated regions in the feature learning phase. Besides, these target-aware features are also more advantageous for subsequent correlation matching to perform a more accurate matching result.

Specifically, we designed a unified pipeline framework based on target-aware attention for CISDL. We adopt the target-aware attention mechanism to naturally integrate feature extraction and correlation matching together. Different from multi-head attention \cite{vaswani2017attention}, target-aware attention takes the two input image features as input and divides the attention process into two heads. One head performs self-attention to extract features from an image, while the other head performs cross-attention to calculate correlation matching and extract features from the other image. Those features from different heads will be concatenated on the channel dimension and mixed by a Mix Feed Forward Network (Mix-FFN). 
In this way, the feature extraction process will be continually adjusted with the help of correlation matching, and these target-aware features will be more beneficial for the subsequent correlation matching.

Apart from the limitation of using a separate pipeline, we also observe that existing methods only consider the matching of image patches at the same scale in their correlation matching module. However, in real-world scenarios, the spliced region is usually processed by scale transformation. Therefore, we further propose a multi-scale projection mechanism to generate image patch tokens with multi-scale information. We integrate this mechanism into each target-aware Attention, naturally enabling the model to perform multi-scale attention mechanism to model the correlation between image patches of different scales. 

Our main contributions can be summarized as follows:

\begin{itemize}
    \item We propose a unified pipeline framework for CISDL. In contrast to previous methods, our method based on target-aware attention jointly learns features and performs correlation matching between the probe and donor images. This approach adaptively reveals the relationship between related image patches for collaborative learning and facilitates mutual promotion of feature learning and correlation matching.  
    \item Considering that spliced region is usually processed by scale transformations, we further design a multi-scale projection method, which can be readily integrated into the attention framework to model the correlation matching between image patches of different scales, which makes our framework more robust for scale transformations.
    \item Experimental results demonstrate the superiority of our proposed model compared with state-of-the-art methods on benchmark datasets.
\end{itemize}

\begin{figure*}[t!]
	\centering
	\includegraphics[width=0.78\linewidth]{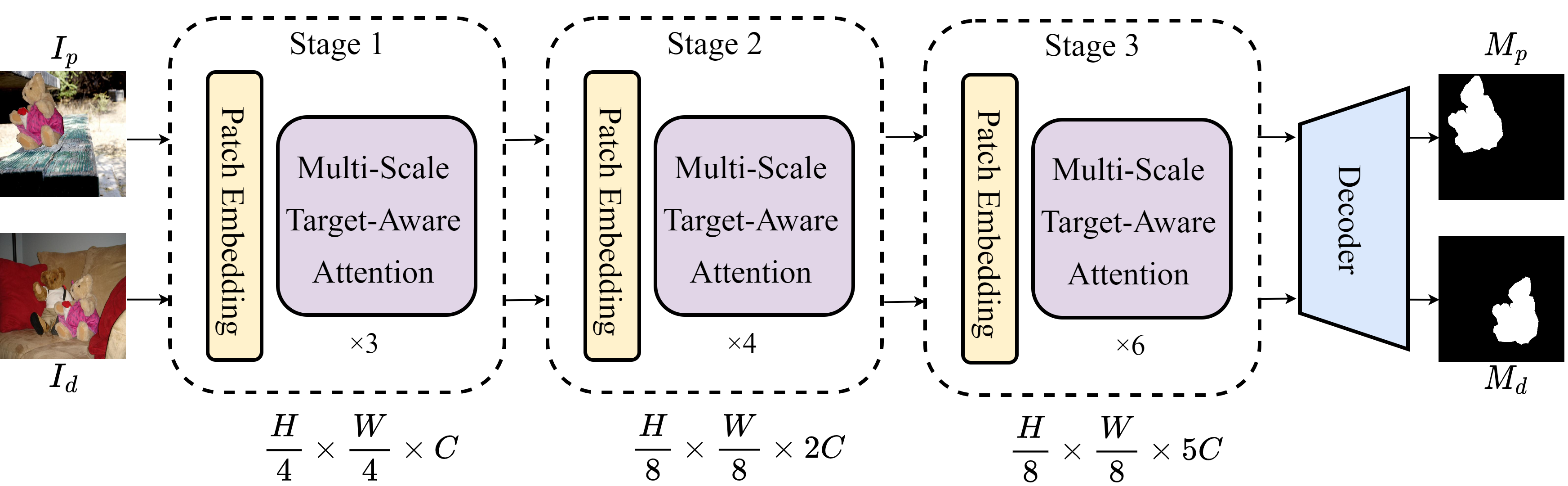}
	\centering
	\caption{The framework of the Multi-scale Target-Aware Framework(MSTAF). The two input images $I_{p}$ and $I_{d}$ are processed by several Patch Embedding and Multi-scale Target-Aware Attention. Then the output feature maps are directly fed into the mask decoder to generate the final masks $M_{p}$ and $M_{d}$. }
	\label{fig:framework}
\end{figure*}

\section{Related Works}
In this section, we will provide a brief overview of some related works concerning existing constrained image splicing detection and localization (CISDL) algorithms as well as the Attention mechanism.

\subsection{CISDL}
In contrast to conventional forgery detection tasks, Constrained Image Splicing Detection and Localization (CISDL) is a new forensic task aiming to identify both the forged and the corresponding source regions in the probe and donor image, respectively. Existing algorithms typically treat CISDL as a matching problem by comparing the similarity of image patches between the two images, similar to copy-move detection. However, traditional copy-move algorithms were designed to detect near-duplicated regions within a single image, which cannot be directly applied to CISDL. 

To address this issue, Wu \textit{et al.} \cite{wu2017deep} proposed the first end-to-end optimized deep matching and validation network (DMVN) for CISDL. DMVN adopts VGG \cite{vgg} as feature extractor and compares high-level features to calculate correlation maps for splicing localization. Ye \textit{et al.} \cite{ye2018feature} proposed a feature pyramid deep matching and localization network (FPLN), which fuses different resolution feature maps from VGG together for subsequent correlation calculations. Liu \textit{et al.} \cite{liu2019adversarial} proposed a deep matching network named DMAC, which removes the pooling layer and adopts atrous convolution in the last VGG block to generate high-resolution feature maps. The methods as mentioned above are mainly aimed at designing a better feature extractor while using simple dot-product calculation for correlation matching still limits the matching performance of the model. To address this problem, Liu \textit{et al.} \cite{liu2020constrained} proposed an attention-aware mechanism to conduct a weighted dot-product in the correlation computation module. Xu \textit{et al.} \cite {xu2022scale} proposed an attention-based separable convolutional block to refine the correlation score maps for generating fine-grained mask prediction. 

\subsection{Attention Mechanism}
There has been a long line of prior research on utilizing the attention mechanism in neural networks. Bahdanau \textit{et al.} \cite{bahdanau2014neural} developed an attention mechanism for machine translation that models dependencies between different parts of the input or output sequences, irrespective of their distance. Since then, numerous subsequent attention-based algorithms for natural language processing have been introduced, such as \cite{luong2015effective,vaswani2017attention}. 
The Transformer, a simple network architecture based exclusively on self-attention mechanisms, was recently proposed to facilitate the parallelization of the training process \cite{vaswani2017attention}. 
With the success of this architecture in machine translation, researchers have begun to apply Transformers to computer vision tasks. The Vision Transformer (ViT) \cite{Dosovitskiy2020} is groundbreaking work on utilizing a pure Transformer encoder for an image classification task. ViT transforms the input image into a sequence of tokens and then inputs them into a standard Transformer encoder with multi-head self-attention. Inspired by this approach, many other Transformer-based vision architectures have been developed, such as PVT \cite{wang2022pvt}, DeiT \cite{touvron2020DeiT}, Swin Transformer \cite{liu2021swin}, Mixformer \cite{cui2022mixformer} and SimTrack \cite{chen2022backbone}.

In this work, we propose a unified attention mechanism, which fully exploits the interaction between the related image patches. Different from the standard attention mechanism, it introduces information interaction into the feature extraction process in a natural way, enabling the model to learn target-aware features.

\section{Proposed Method}
As shown in Fig.\ref{fig:framework}, our framework contains several Overlap Patch Embedding modules and Multi-scale Target-Aware Attention modules. After the final Multi-scale Target-Aware Attention modules, the features are directly sent to a decoder that generates masks indicating the spliced region. Additionally, we further designed a multi-scale attention mechanism to model the matching between image patch tokens of different scales. Details of the proposed framework are described in the following subsections.

\subsection{Overlap Patch Embedding}
Given a probe image $I_p$ and a donor image $I_d$ as input, we first embed them into visual representation of image patches. To preserve the local continuity around image patches and perform finer-grained matching, we use an overlap patch embedding process, which can be implemented by a convolutional layer and a layer normalization layer. As shown in Fig. \ref{fig:framework}, each Overlap Patch Embedding layer embeds the image patches into tokens with different resolutions and divides the framework into three different stages. Let's take stage 1 as an example to illustrate the processing flow of the model. For example, the first overlapping patch embedding layer takes the $I_p\in{\mathbb{R}^{H\times W\times 3}}$ and $I_d\in{\mathbb{R}^{H\times W\times 3}}$ as input. It uses a $7\times7$ convolutional layer with a stride of 4 to divide the two images into overlapping image patch tokens. Then we use layer normalization to process $F_p$ and $F_d$ and flatten them for subsequent dual attention computation. This process can be represented by the following formula.

\begin{eqnarray}
F_i = Flatten(LN(Conv(I_i)))\quad i\in\{p, d\}
\end{eqnarray}

\noindent where $F_i\in{\mathbb{R}^{N \times C}}$. N is the number of image patch tokens and is equal to $\frac{H}{4} \times \frac{W}{4}$.
\begin{figure}[t!]
	\centering
	\includegraphics[width=1.0\linewidth]{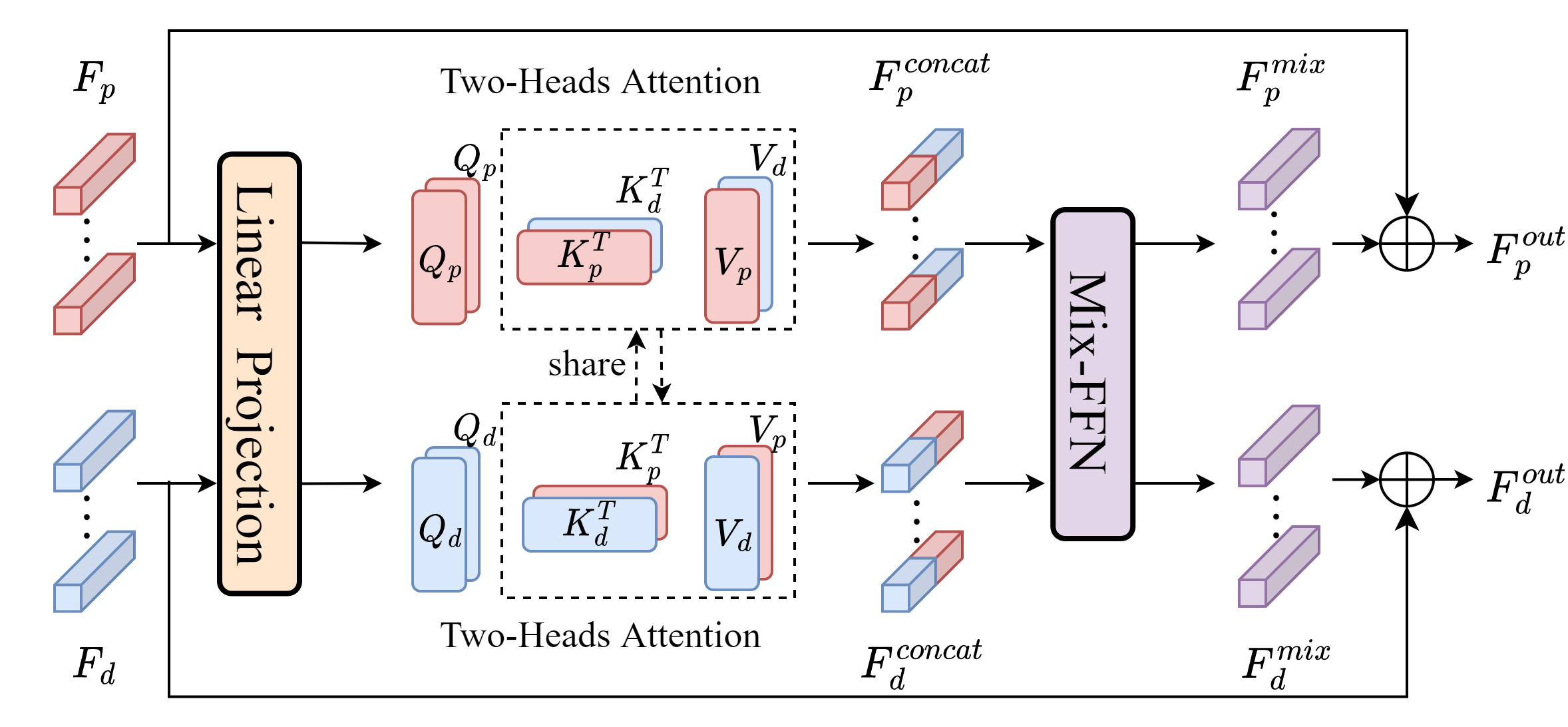}
	\centering
	\caption{Target-Aware Attention. The features of probe image are represented in red, while the features of donor image are represented in blue. The purple color means that after processing by Mix-FFN module, features from different images are fused together. }
	\label{fig:Dual Attention}
\end{figure}

\subsection{Target-Aware Attention}
Inspired by \cite{cui2022mixformer, chen2022backbone}, we design target-aware attention to implement feature extraction and correlation matching jointly, as shown in Fig.\ref{fig:Dual Attention}. Different from them, our target-aware attention utilizes two distinct attention heads to perform feature extraction and correlation matching separately, which avoids interference between self-attention and cross-attention. With the help of target-aware attention, our unified pipeline model can greatly promote collaborative learning of the related regions in the two images and suppress the effects from the background and other unrelated regions in the feature learning phase.

Specifically, we first use a linear projection layer to project $F_i$ into $Q_i, K_i, V_i$ respectively, which can be represented by the following formula:

\begin{eqnarray}
Q_i = (F_i)^{T}W_Q \quad K_i = (F_i)^{T}W_K \quad V_i = (F_i)^{T}W_V
\end{eqnarray}

\noindent where $i\in\{p, d\}$, $Q_i,K_i,V_i\in{\mathbb{R}^{N \times \frac{C}{2}}}$ and $W_Q,W_K,W_V$ represent linear projection. Then we perform Target-Attention on two attention heads respectively:

\begin{eqnarray}
head_{self} = Softmax(\frac{Q_iK_i^T}{\sqrt{C}})V_i\quad i\in\{p, d\}
\end{eqnarray}

\begin{eqnarray}
head_{cross} = Softmax(\frac{Q_iK_j^T}{\sqrt{C}})V_j\quad i,j\in\{p, d\} \quad i \neq j
\end{eqnarray}

\noindent where $C$ is the dimension of the image patch token. In $head_{self}$, $Q, K, V$ are from the same image. It means that $head_{self}$ is responsible for extracting features from the input image itself. In $head_{cross}$, $Q$ and $K, V$ are from different images, which means that $head_{cross}$ is responsible to perform correlation matching and extracting features from the other image. $Q_iK_j^T$ represents the correlation matching process between the two input images, which is an affinity matrix that indicates the correlation of each image patch pair between the two input images. Multiplying $Q_iK_j^T$ with $V_j$ represents selecting and extracting the high-related features from the other image. Then we concatenate the two sets of features from the two different heads on the channel dimension.

\begin{eqnarray}
Attention = Concat(head_{self}, head_{cross})
\end{eqnarray}

After target-aware attention, we obtain $F^{concat}_i\in{\mathbb{R}^{N \times C}}$. As shown in Fig. \ref{fig:Dual Attention}, The first $\frac{C}{2}$ channels of $F^{concat}_i$ represent the features extracted from the input image itself, while the last $\frac{C}{2}$ channels represent the highly-related features selected from the other image using correlation matching. For the spliced region, these two sets of features will be similar, while the background corresponds to two dissimilar sets of features.  With the help of the unified pipeline, our model can perform mutual promotion for feature extraction and correlation matching. The intermediate features of correlation matching can greatly enable the collaborative learning of related regions and suppress the influence of background and other unrelated regions during feature learning.

Subsequently, we feed $F^{concat}_i$ into the Mix-FFN module, which is a feed-forward network (FFN) to mix the two sets of features together. Inspired by \cite{xie2021segformer}, we add a ${3\times3}$ depth-wise convolutional layer to provide local continuity and positional information. Mix-FFN can be formulated as:

\begin{eqnarray}
F^{mix}_i = MLP(GELU(Conv_{3\times3}(MLP(F^{concat}_i))) \quad i\in\{p, d\}
\end{eqnarray}

After being processed by the Mix-FFN module, features from the two heads are fused together. The features of the spliced region will be more discriminative because of fusing similar features from the two images, while the background features will be suppressed because the other image does not contain similar features. 

Finally, we use a residual structure to obtain the output of target-aware attention.

\begin{eqnarray}
F^{out}_i = F_i + F^{mix}_i \quad i\in\{p, d\}
\end{eqnarray}

The $F^{out}_i$ are the input for the next Target-Aware Attention module. Each stage contains several Target-Aware Attention modules, making this unified structure throughout the entire learning phase.

\begin{figure}[t!]
	\centering
	\includegraphics[width=1.0\linewidth]{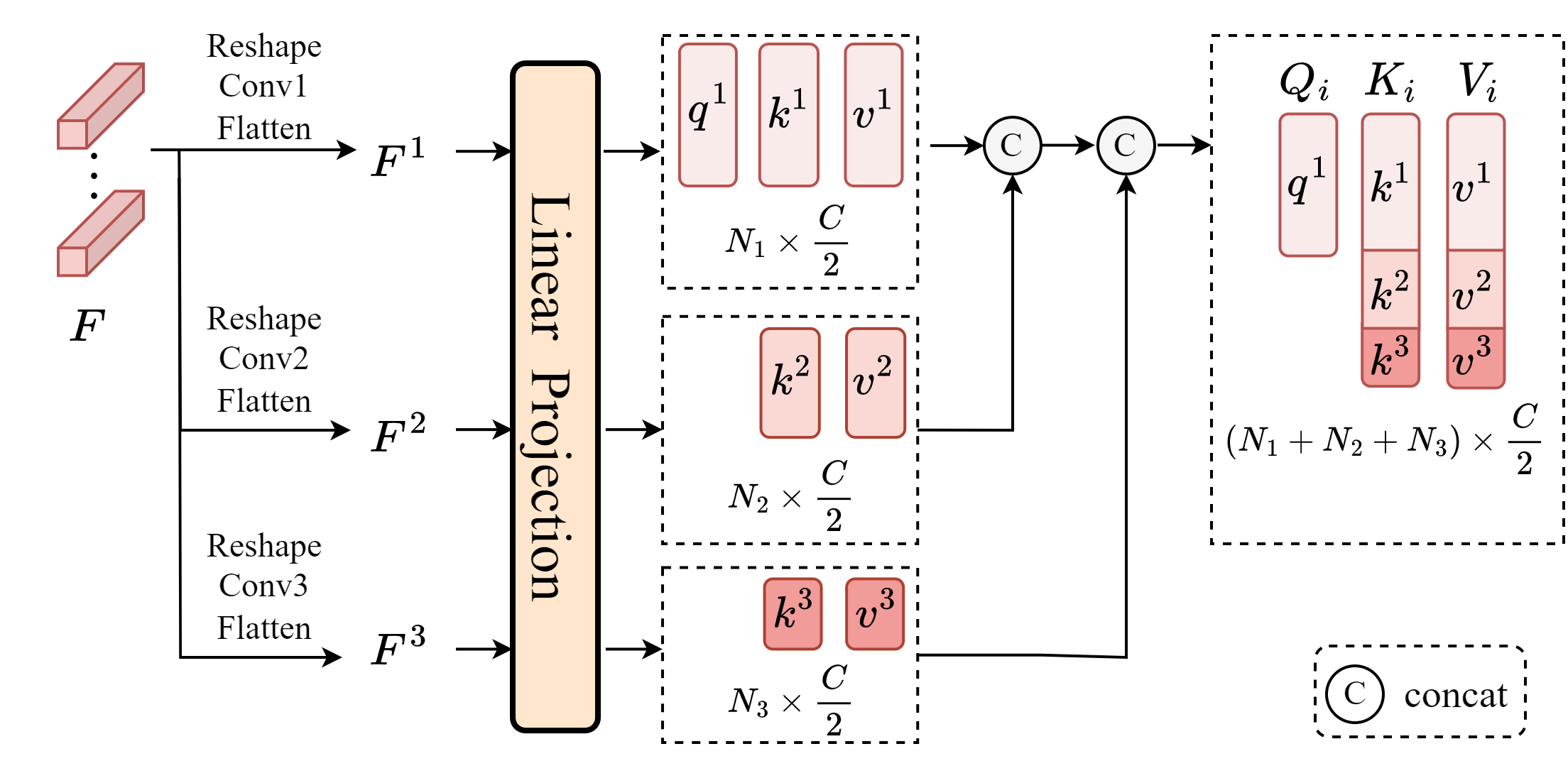}
	\centering
	\caption{Multi-scale Projection Mechanism. Where $i\in\{p, d\}$. The feature maps are processed by two different convolutional layers to obtain features of different scales.  }
	\label{fig: multi}
\end{figure}

\subsection{Multi-scale Projection Mechanism}
The attention mentioned above only performs matching between image patches of the same scale. However, the spliced regions between two images usually have different sizes because of scale transformations of different degrees. Therefore, we further design a multi-scale projection mechanism. As shown in Fig.\ref{fig: multi}, we reshape the $F$ and feed them into different convolutional layers to obtain three sets of feature maps $F^1,F^2,F^3$. $F$ can be $F_p$ or $F_d$. Each convolutional layer has different kernel size and stride to obtain features of different scales. After the convolutional layer, image patch tokens from $F^1,F^2,F^3$ have different receptive fields, which contain different scale image patch information. Then $F^1,F^2,F^3\in{\mathbb{R}^{N_i\times \frac{C}{2}}}$ are processed by a linear projection to obtain $q^i, k^i, v^i\in{\mathbb{R}^{N_i\times \frac{C}{2}}}$. 
For $F^2,F^3$, we only perform $k, v$ projection to reduce spatial redundancy. Finally, all the $q^i, k^i, v^i$ are concatenated together to obtain the final $Q^i\in{\mathbb{R}^{N_1\times \frac{C}{2}}}$ and $K^i,V^i\in{\mathbb{R}^{{(N_1+N_2+N_3)}\times \frac{C}{2}}}$. $N_1, N_2, N_3$ are the numbers of image patch tokens from the three feature maps $F^1,F^2,F^3$ respectively.

We replace the Linear Projection in each Target-Aware Attention module with the Multi-scale projection Mechanism, which can provide $K_i, V_i$ containing tokens of different scales. In this way, we can perform multi-scale attention process to model the matching between image patches of different scales.

\subsection{Mask Decoder and Loss Function}
Existing methods use correlation maps to localize the spliced region, which lacks spatial information. Since the Target-Aware Attention module outputs discriminative features, we directly feed them into a fully-convolutional decoder to generate the final masks. Our decoder consists of several upsampling and convolutional layers to gradually increase the resolution of feature maps by a factor of 2. The final predicted masks have the same size as input images. We use binary cross entropy (BCE) loss and calculate the average loss of the two output masks: 

\begin{eqnarray}
Loss = -\frac{1}{HW} \sum^{HW}_{i}G^{(i)}log(M^{(i)}) + (1-G^{(i)})log(1-M^{(i)})
\end{eqnarray}

\noindent where $G^{(i)}$ represents the pixel in ground truth, where 1 means the spliced pixel and 0 represents the background pixel. $M^{(i)}$ is pixel in the prediction mask. Its value ranges between 0 and 1, representing the probability that this pixel belongs to the spliced region.

\begin{table*}[t]
\begin{center}
\caption{Comparison on the Synthetic set} \label{tab:Synthetic}
\begin{tabular}{c|c c c|c c c|c c c}
  \hline
  \multirow{2}*{Method} & \multicolumn{3}{|c|}{Difficult} & \multicolumn{3}{|c|}{Normal}  & \multicolumn{3}{|c}{Easy}
  \\ \cline{2-10}
 & IoU & MCC & NMM & IoU & MCC & NMM & IoU & MCC & NMM 
  \\ \hline
  DMVN \cite{wu2017deep} & 0.2772 & 0.3533 & -0.4382 & 0.6818 & 0.7570 & 0.4042 & 0.8198 & 0.8544 & 0.6770
  \\ 
  DMAC-adv \cite{liu2019adversarial} & 0.5433 & 0.6584 & 0.1026 & 0.8317 & 0.8833 & 0.6877 & 0.9237 & 0.9411 & 0.8655
  \\  
  AttentionDM \cite{liu2020constrained} & 0.7228 & 0.8108 & 0.4793 & 0.8980 & 0.9320 & 0.8253 & 0.9602 & 0.9603 & 0.9388
  \\  
  SADM \cite{xu2022scale} & 0.7759 & 0.8128 & 0.5129 & 0.9040 & 0.8288 & 0.8265 & 0.9621 & 0.9616 & 0.9410
  \\  \hline
  MSTAF (ours) & \textbf{0.8394} & \textbf{0.8918} & \textbf{0.7064} & \textbf{0.9510} & \textbf{0.9700} & \textbf{0.9151} & \textbf{0.9788} & \textbf{0.9838} & \textbf{0.9646} \\ \hline
\end{tabular}
\end{center}
\end{table*}

\begin{table*}[t]
\begin{center}
\caption{Comparison on the MFC2018 dataset} \label{tab:MFC2018}
\begin{tabular}{c|c c c|c c c}
  \hline
  \multirow{2}*{Method} & \multicolumn{3}{|c|}{Localization} & \multicolumn{3}{|c}{Detection} 
  \\ \cline{2-7}
 & IoU & MCC & NMM & Precision & Recall & F1-score 
  \\ \hline
  DMVN \cite{wu2017deep} & 0.1204 & 0.1560 & -0.7710 & \textbf{0.9099} & 0.3594 & 0.5153
  \\ 
  DMAC-adv \cite{liu2019adversarial} & 0.2071 & 0.2484 & -0.6120 & 0.7162 & 0.5836 & 0.6431 
  \\  \hline
  MSTAF (ours) & \textbf{0.3576} & \textbf{0.3934} & \textbf{-0.3037} & 0.8868 & \textbf{0.6595} & \textbf{0.7565} 
  \\ \hline
\end{tabular}
\end{center}
\end{table*}

\begin{figure*}[t!]
	\centering
	\includegraphics[width=0.65\linewidth]{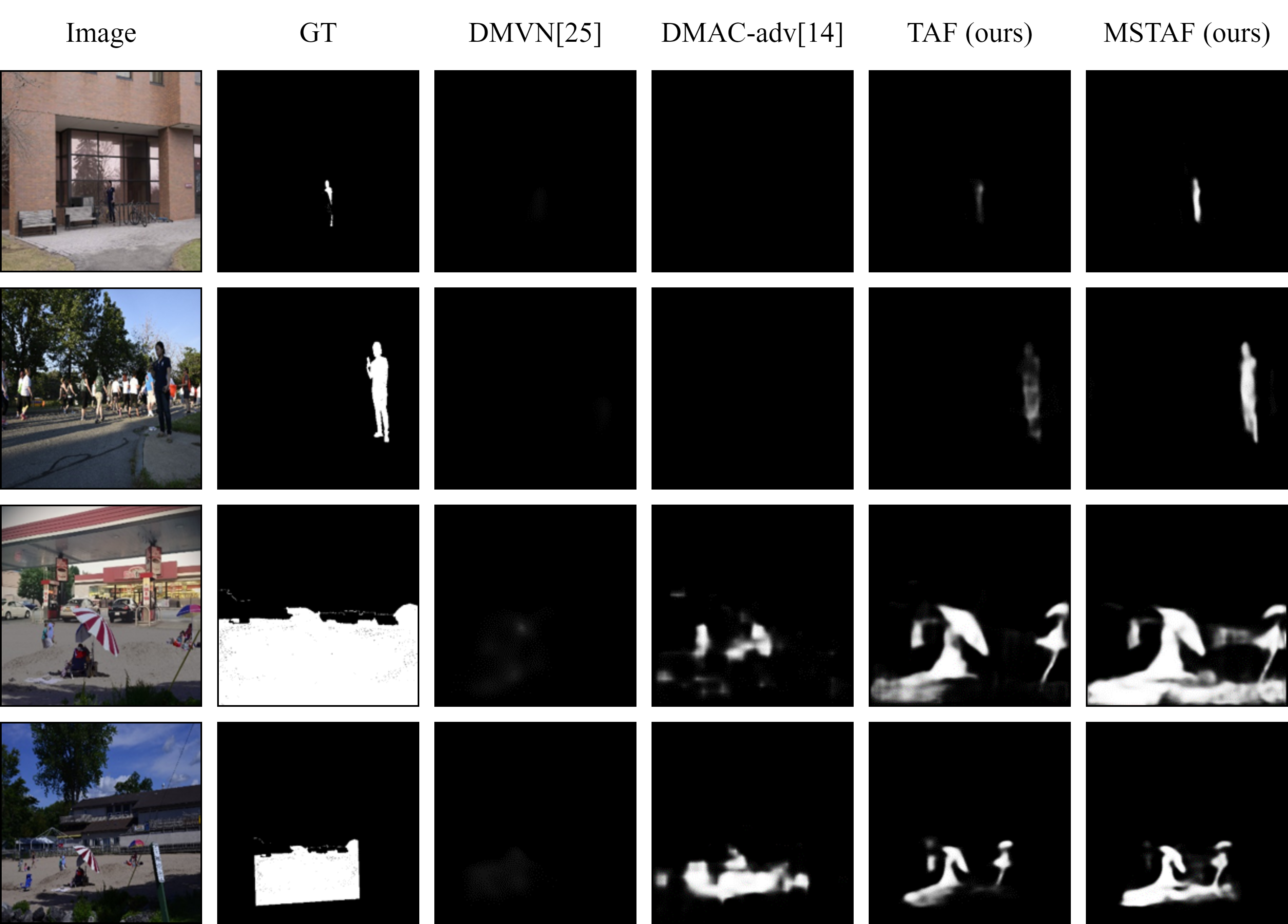}
	\centering
	\caption{Visualization of localization performance on MFC2018. GT is ground truth. TAF is our model without multi-scale projection mechanism.}
	\label{fig:MFC2018}
\end{figure*}

\begin{table}[t]
\begin{center}
\caption{Comparison on the paired CASIA dataset} \label{tab:casia}
\begin{tabular}{c|c c c}
  \hline
   Method & Precision & Recall & F1-score
  \\ \hline
  Chrislein \textit{et al.} \cite{Christlein} & 0.5164 & 0.8292 & 0.6364 
  \\
  Luo \textit{et al.} \cite{Luo} & \textbf{0.9969} & 0.5353 & 0.6966  
  \\
  Ryu \textit{et al.} \cite{Ryu} & 0.9614 & 0.5895 & 0.7309 
  \\
  Cozzolino \textit{et al.} \cite{Cozzolino} & 0.9897 & 0.6334 & 0.7725 
  \\ \hline
  DMVN-loc \cite{wu2017deep} & 0.9152 & 0.7918 & 0.8491 
  \\ 
  DMVN-det \cite{wu2017deep} & 0.9415 & 0.7908 & 0.8596 
  \\ 
  DMAC-adv \cite{liu2019adversarial} & 0.9657 & 0.8576 & 0.9085
  \\ 
  AttentionDM \cite{liu2020constrained} & 0.9288 & 0.9204 & 0.9246
  \\
  SADM \cite{xu2022scale} & 0.9741 & 0.8758 & 0.9263
  \\\hline
  MSTAF (ours)  & 0.9815 & \textbf{0.9251} & \textbf{0.9524}\\ \hline
\end{tabular}
\end{center}
\end{table}

\begin{table*}[t]
\begin{center}
\caption{Ablation study on the Synthetic set} \label{tab:ablation1}
\begin{tabular}{c|c c c|c c c|c c c}
  \hline
  \multirow{2}*{Model} & \multicolumn{3}{|c|}{Difficult} & \multicolumn{3}{|c|}{Normal}  & \multicolumn{3}{|c}{Easy}
  \\ \cline{2-10}
    & IoU & MCC & NMM & IoU & MCC & NMM & IoU & MCC & NMM 
  \\ \hline
  TAF-separate & 0.6239 & 0.7167 & 0.2963 & 0.8738 & 0.9143 & 0.7744 & 0.9490 & 0.9604 & 0.9164
  \\ 
  MSTAF-separate  & 0.7712 & 0.8432 & 0.5670 & 0.9210 & 0.9486 & 0.8583 & 0.9693 & 0.9764 & 0.9488
  \\  \hline
  TAF  & 0.8001 & 0.8586 & 0.6312 & 0.9379 & 0.9605 & 0.8937 & 0.9753 & 0.9811 & 0.9617
  \\  
  MSTAF & \textbf{0.8394} & \textbf{0.8918} & \textbf{0.7064} & \textbf{0.9510} & \textbf{0.9700} & \textbf{0.9151} & \textbf{0.9788} & \textbf{0.9838} & \textbf{0.9646} \\ \hline
\end{tabular}
\end{center}
\end{table*}

\begin{table*}[t]
\begin{center}
\caption{Ablation study on the Scale set} \label{tab:ablation2}
\begin{tabular}{c|c c c|c c c|c c c}
  \hline
  \multirow{2}*{Model}  & \multicolumn{3}{|c|}{Difficult} & \multicolumn{3}{|c|}{Normal}  & \multicolumn{3}{|c}{Easy}
  \\ \cline{2-10}& IoU & MCC & NMM & IoU & MCC & NMM & IoU & MCC & NMM 
  \\ \hline
  TAF  & 0.7009 & 0.7644 & 0.4400 & 0.8789 & 0.9160 & 0.7807 & 0.9704 & 0.9769 & 0.9550
  \\  
  MSTAF & \textbf{0.7427} & \textbf{0.8022} & \textbf{0.5206} & \textbf{0.9105} & \textbf{0.9410} & \textbf{0.8406} & \textbf{0.9752} & \textbf{0.9808} & \textbf{0.9602} \\ \hline
\end{tabular}
\end{center}
\end{table*}

\section{EXPERIMENT}

\subsection{Training Data and Implementation Details}
We use the training subset of MS COCO dataset \cite{lin2014coco} to generate training pairs according to the generation scripts provided by \cite{liu2019adversarial}. The generation process randomly segments an object from an image by the annotation and splices it into another randomly selected image to generate an image pair. The spliced object are processed by five types of transformation, i.e., shift, rotation, scale, luminance, and deformation. For fare comparison, we generated 1M (Million) training pairs and resized them to 256 $\times$ 256, which is consistent with other CISDL methods. We trained our model for 6 epochs and the batch size is set to 10. The optimizer is Adam and the learning rate is set to 1e-4. We implemented our proposed method on PyTorch and train our model on NVIDIA 3090 GPU. The number of Target-Aware Attention modules is set to 3, 4, 6 in each stage respectively and the embedding dimension C is set to 64. 

\subsection{Test Dataset and Evaluation Metric}
We evaluate localization (pixel-level) performance and detection (image-level) performance on three datasets:(1)The Synthetic set. (2)The MFC2018 dataset. (3)The paired CASIA dataset. To be consistent with \cite{xu2022scale}, We adopted IoU (Intersection over Union), NMM (Nimble Mask Metric), and MCC (Matthews Correlation Coefficient) to evaluate the localization performance. The detection evaluation metrics are Precision, Recall, and F1. The detection strategy is followed as \cite{wu2017deep}. We first process the two predicted masks with a binary threshold of 0.5. Then we determined whether an image pair is positive by checking if there is any pixel equal to 1 in the two processed masks. 

The Synthetic set. In \cite{liu2019adversarial}, they generated 9000 testing image pairs by the same process described in 4.1. These testing image pairs are generated from testing subsets of MS COCO without overlapping with training data. The spliced regions are processed by several kinds of transformations. The whole dataset is equally divided into Difficult, Normal, Easy sets according to the proportions of the spliced region.

The MFC2018 dataset. The Media Forensics Challenge 2018 dataset \cite{NISTMediaForensics2018} is a challenging dataset. It provides 1327 positive image pairs and 16673 negative image pairs with corresponding ground truth masks. We found that this dataset contains some samples where the proportion of spliced pixels was less than 1\%. We removed these extremely difficult samples and the number of positive image pairs for evaluation was reduced to 843. We use these positive image pairs to evaluate the localization performance. We also use the whole dataset (843 positive image pairs and 16673 negative image pairs) to evaluate the detection performance. 

The paired CASIA dataset. In \cite{wu2017deep}, they selected 3642 positive pairs and 5000 negative pairs from CASIAv2.0 \cite{dong2013casia} to create this new dataset for CISDL problem. We can only perform detection evaluation and conduct quantitative comparison on this dataset due to the lack of ground truth masks.

\begin{figure*}[t!]
	\centering
	\includegraphics[width=0.65\linewidth]{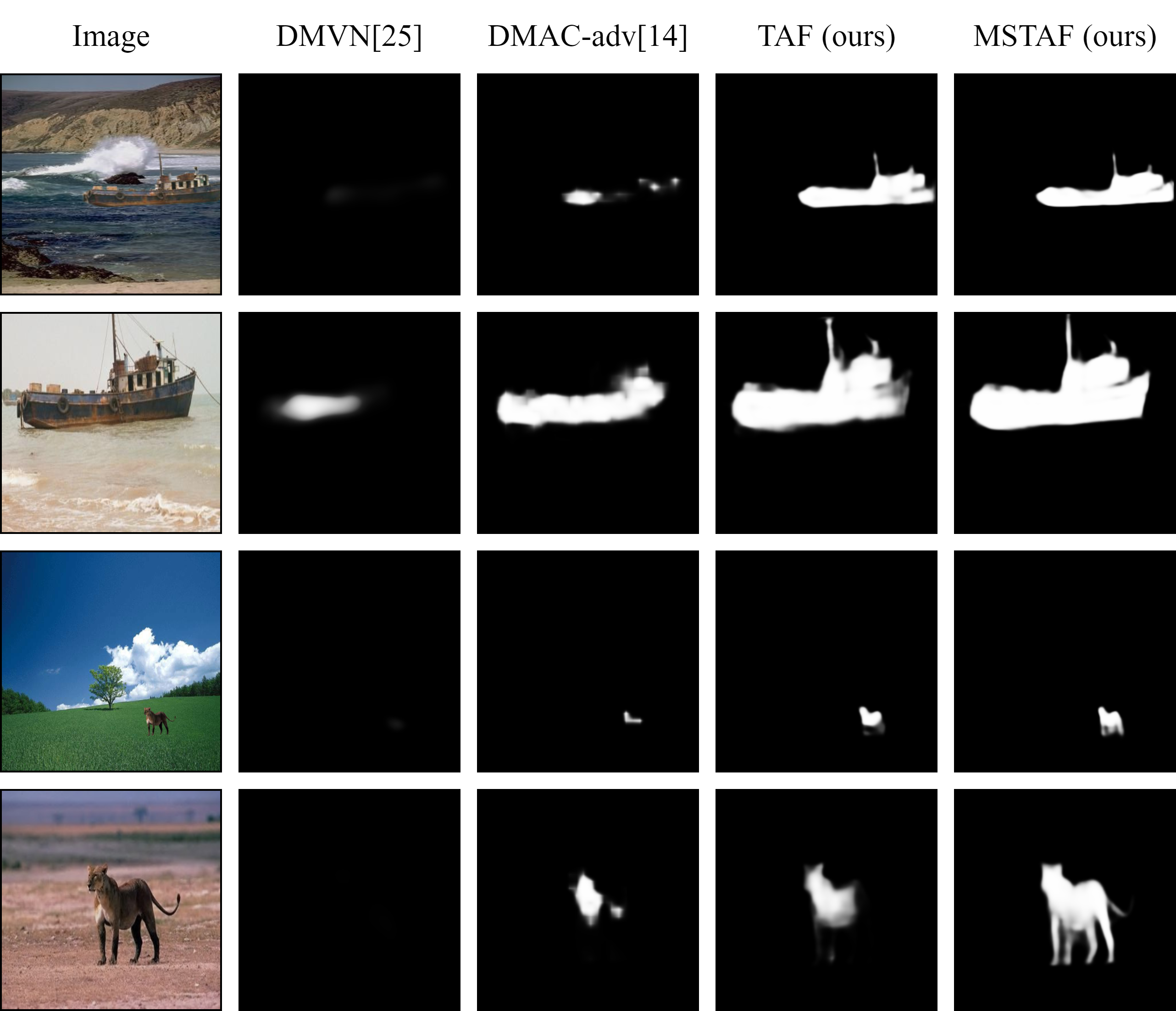}
	\centering
	\caption{Visualization of the paired CASIA dataset. The paired CASIA dataset does not provide ground truth. TAF is our model without multi-scale projection mechanism.}
	\label{fig:CASIA}
\end{figure*}

\subsection{Comparison Result}
 We compared the localization performance with current methods on the Synthetic set. As shown in Tab \ref{tab:Synthetic}, our Multi-scale Target-Aware Framework (MSTAF) outperforms state-of-the-art methods under all evaluation metrics on all subsets. In Difficult set, the proportions of spliced regions are only 1\%-10\% of the whole image, which is challenging for localization. However, our method has the most significant improvement on the difficult set, demonstrating that our method greatly improves the ability to localize small regions. 

 We both evaluate the localization and detection performance on MFC2018 dataset. As shown in Tab \ref{tab:MFC2018}, MSTAF achieved the best localization performance compared to DMVN \cite{wu2017deep} and DMAC-adv \cite{liu2019adversarial}. As AttentionDM \cite{liu2020constrained} and SADM \cite{xu2022scale} have not been provided their code, we cannot obtain their models and compare our method with them on this dataset. We present the visualization of localization performance in Fig \ref{fig:MFC2018}. TAF is our model without multi-scale projection mechanism. We can see that MFC2018 is a challenging dataset, which contains many crafted manipulated samples that have been processed by scale and rotation transformations. While DMVN \cite{wu2017deep} and DMAC-adv \cite{liu2019adversarial} fail to localize the splicing region of different scales, our methods can accurately localize the spliced regions. Besides, with the help of the multi-scale projection mechanism, MSTAF is more robust against scale transformation and presents better localization performance than TAF. 

The detection performance comparison on the paired CASIA dataset is shown in Tab \ref{tab:casia}. In \cite{wu2017deep}, they compared with several copy-move methods\cite{Christlein,Luo,Ryu,Cozzolino} due to the lack of other CISDL methods. We borrowed their evaluation scores and compared them with current CISDL methods. Although the method proposed in \cite{Luo} achieved the highest precision, its recall was the lowest because it classifies many negative samples as positive. It can see that our model achieves the highest F1 score, which indicates that our model can achieve the best trade-off between precision and recall. Visual comparisons as shown in Fig \ref{fig:CASIA}. We can see that our method can provide more accurate localization performance. Besides, the effectiveness of DMVN and DMAC diminishes significantly when the spliced region is processed by scale transformation. However, MSTAF still achieves great localization performance, which demonstrates its robustness against scale transformation.

\subsection{Visualization}
\begin{figure}[t!]
	\centering
	\includegraphics[width=0.9\linewidth]{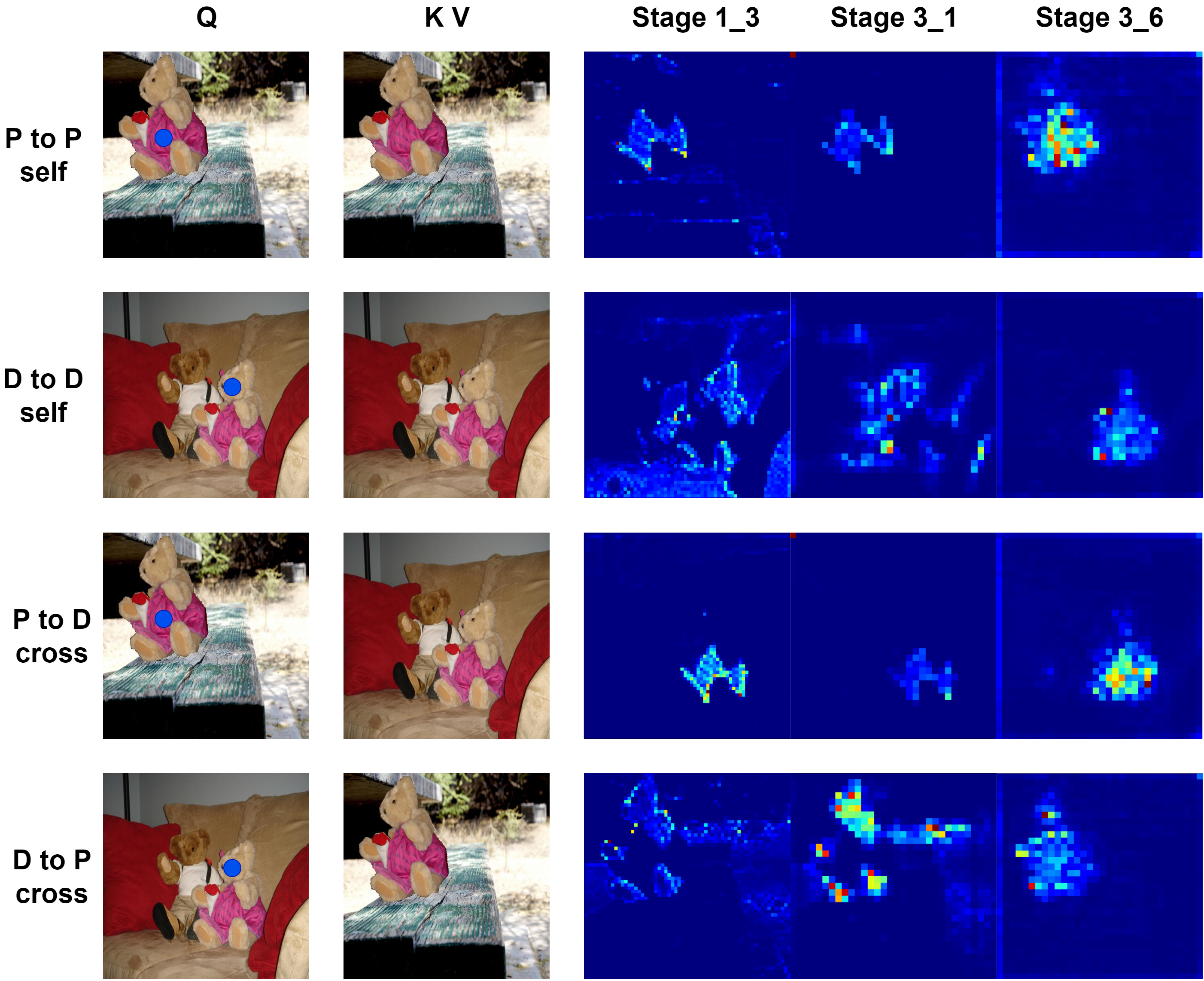}
	\centering
	\caption{Visualization of attention maps. The blue point represents the token we selected to present attention maps. Stage i\_j represents attention maps from the j Target-Aware Attention modules in stage i.}
	\label{fig:Attention}
\end{figure}

To explore the effect of attention in different stages of the model, we visualized some attention maps, which are shown in Fig \ref{fig:Attention}. With flexible and global modeling capabilities, both self-attention and cross-attention can gradually focus on the whole spliced object and provides features with more object-level information. Benefited from cross-attention providing continuous communication between the two images, those irrelevant features are filtered out so that self-attention can extract target-aware features to perform better correlation matching. 

\subsection{Ablation}
We design a unified pipeline implemented by Target-Aware Attention Framework (TAF), which is different from the separate pipeline used by existing methods. In order to evaluate the effectiveness of the unified pipeline, we built a separate pipeline model for comparison. The TAF-separate model adopts the same architecture as TAF. It implements self-attention for both two heads of all Target-Aware Attention modules in the front, while the last Target-Aware Attention module only implements cross-attention for both two heads. We use the Synthetic dataset to conduct ablation study, the results are shown in Table \ref{tab:ablation1}. By comparing the TAF-separate model and the TAF model, we can see that the TAF with unified pipeline performs better than the separate pipeline. After introducing the multi-scale projection mechanism to achieve multi-scale attention, the MSTAF-separate and MSTAF models both showed improvements in localization performance. The MSTAF with unified pipeline design and multi-scale attention mechanism presents the best localization performance on all three subsets.

In \cite{liu2019adversarial}, they use the Scale set to analyze the model's robustness against scale transformation. To further verify the effectiveness of the multi-scale attention mechanism. We also use Scale set to evaluate our models and the result as shown in Tab \ref{tab:ablation2}. In the Scale set, the spliced region is only processed with scale transformation of different degrees for ablation in \cite{liu2019adversarial}. It contains 9000 testing pairs and is equally divided into Difficult, Normal, Easy subsets. In the Difficult subset and Normal subset, there are more samples with larger scale degrees. On the contrary, in the Easy subset, the size of the spliced region between the two images tends to be more consistent. We can see that with the help of multi-scale attention mechanism, MSTAF achieves much better localization performance than TAF on the Normal and difficult subset. It demonstrates that MSTAF has an advantage in dealing with various scale transformation samples. 
After introducing the multi-scale attention mechanism, MSTAF is more robust against scale transformation. The visual comparison can refer to Fig. \ref{fig:MFC2018} and Fig. \ref{fig:CASIA}.

\section{CONCLUSION}
In this work, we propose a Multi-scale Target-Aware Framework to simplify the pipeline of existing methods. It adopts self-attention for feature extraction and cross-attention for correlation matching simultaneously. This unified design enables feature extraction and correlation matching to mutually promote each other, thereby enhancing the matching performance of the model. We further design a multi-scale attention mechanism to model the matching between image patches of different scales, which further improves the robustness against scale transformation. Experiment results demonstrate that our model is robust against scaling and outperforms state-of-the-art methods.

\begin{acks}
This work was supported in part by the Natural Science Foundation of China under Grant 62001304; in part by the Guangdong Basic and Applied Basic Research Foundation under Grant 2022A1515010645; in part by the Foundation for Science and Technology Innovation of Shenzhen under Grant RCBS20210609103708014 and the Key project of Shenzhen Science and Technology Plan under Grant 20220810180617001; in part by CCF-Alibaba Innovative Research Fund For Young Scholars; in part by the Open Research Project Programme of the State Key Laboratory of Internet of Things for Smart City (University of Macau) under Grant SKLIoTSC(UM)-2021-2023/ORP/GA04/2022.
\end{acks}

\bibliographystyle{ACM-Reference-Format}
\balance
\bibliography{acmart}

\end{document}